\documentclass[10pt,twocolumn,letterpaper]{article}

\usepackage{wacv} 
\usepackage{graphicx}
\usepackage{amsmath}
\usepackage{amssymb}
\usepackage{booktabs}
\usepackage{bm}
\usepackage{makecell}
\usepackage{graphbox}
\usepackage{cuted}
\usepackage{caption}
\usepackage{enumitem}
\usepackage{mathtools}
\usepackage{blindtext}
\usepackage[normalem]{ulem}
\usepackage{tikz}

\newcommand{\autourblue}[1]{\tikz[baseline=(X.base)]\node [draw=blue,fill=gray!10,semithick,rectangle,inner sep=2pt, rounded corners=3pt] (X) {#1};}
\newcommand{\autourgreen}[1]{\tikz[baseline=(X.base)]\node [draw=black!60!green,fill=gray!10,semithick,rectangle,inner sep=2pt, rounded corners=3pt] (X) {#1};}
\newcommand{\autourbluedashed}[1]{\tikz[baseline=(X.base)]\node [draw=blue,fill=gray!10,semithick,rectangle,inner sep=2pt, rounded corners=3pt, dashed] (X) {#1};}
\newcommand{\autourgreendashed}[1]{\tikz[baseline=(X.base)]\node [draw=black!60!green,fill=gray!10,semithick,rectangle,inner sep=2pt, rounded corners=3pt, dashed] (X) {#1};}

\usepackage{xfrac}
\usepackage{multirow}
\usepackage{subcaption}
\usepackage{comment}
\usepackage[accsupp]{axessibility}
\usepackage[pagebackref,breaklinks,colorlinks]{hyperref}
\usepackage[capitalize]{cleveref}

\crefname{section}{Sec.}{Secs.}
\Crefname{section}{Section}{Sections}
\Crefname{table}{Table}{Tables}
\crefname{table}{Tab.}{Tabs.}

\begin{document}
\title{Exploiting the Signal-Leak Bias in Diffusion Models}
\author{Martin Nicolas Everaert$^1$\qquad Athanasios Fitsios$^{1, 2}$\qquad Marco Bocchio$^2$\\Sami Arpa$^2$\qquad Sabine S{ü}sstrunk$^1$\qquad Radhakrishna Achanta$^1$\\
\normalsize $^1$School of Computer and Communication Sciences, EPFL, Switzerland \quad $^2$\href{https://home.largo.ai/}{Largo.ai}, Lausanne, Switzerland
\\
\normalsize Project page: \url{https://ivrl.github.io/signal-leak-bias/}
}
\maketitle

\begin{strip}
    \centering
    \setlength{\tabcolsep}{1pt}
    \begin{tabular}{p{1.5cm}p{2.6cm}p{1.5cm}|p{1.7cm}p{2.6cm}p{1.5cm}|p{1.5cm}p{2.6cm}p{1.5cm}}
    
        \multicolumn{3}{c|}{\makecell[c]{
            Data distribution \autourblue{$q(\bm{x}_{0})$} approximately \\
            matches noise distribution \autourbluedashed{$p_\text{noise}$}
        }} & 
        \multicolumn{3}{c|}{\makecell[c]{
            Data distribution \autourblue{$q(\bm{x}_{0})$} does not \\
            match noise distribution \autourbluedashed{$p_\text{noise}$}, \\
            creating a significant signal leakage
        }} & 
        \multicolumn{3}{c}{\makecell[c]{
            Our proposed correction accounting \\ for the signal leakage and realigning  \\ \autourgreendashed{$q(\hat{\bm{x}}_{T})$} with \autourgreen{$q(\bm{x}_{T})$}}
        } \\
        
        \multicolumn{3}{c|}{\includegraphics[align=c, width=.31\linewidth]{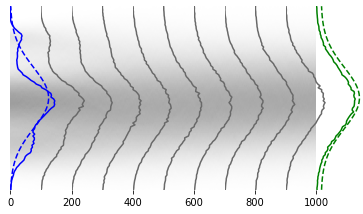}} & 
        \multicolumn{3}{c|}{\includegraphics[align=c, width=.31\linewidth]{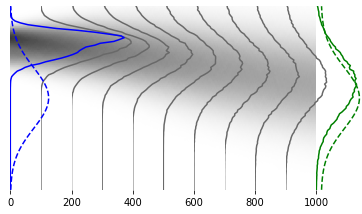}} & 
        \multicolumn{3}{c}{\includegraphics[align=c, width=.31\linewidth]{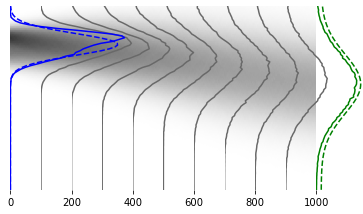}} \\

        \makecell[c]{\autourblue{$q(\bm{x}_{0})$} \\ \autourbluedashed{$p_\text{noise}$}} & 
        \makecell[c]{\small{Noising process $\rightarrow$}\\ $q(\bm{x}_{t}), t \in [0, T]$\vspace{2mm}} & 
        \makecell[c]{\autourgreen{$q(\bm{x}_{T})$} \\ \autourgreendashed{$q(\hat{\bm{x}}_{T})$}} &
        
        \makecell[c]{\autourblue{$q(\bm{x}_{0})$} \\ \autourbluedashed{$p_\text{noise}$}} & 
        \makecell[c]{\small{Noising process $\rightarrow$}\\$q(\bm{x}_{t}), t \in [0, T]$\vspace{2mm}} & 
        \makecell[c]{\autourgreen{$q(\bm{x}_{T})$} \\ \autourgreendashed{$q(\hat{\bm{x}}_{T})$}} &
        
        \makecell[c]{\autourblue{$q(\bm{x}_{0})$} \\ \autourbluedashed{$\tilde{q_0}$}} & 
        \makecell[c]{\small{Noising process $\rightarrow$}\\$q(\bm{x}_{t}), t \in [0, T]$\vspace{2mm}} & 
        \makecell[c]{\autourgreen{$q(\bm{x}_{T})$} \\ \autourgreendashed{$q(\hat{\bm{x}}_{T})$}} \\[-3pt]
    \end{tabular}
   \captionof{figure}{Current diffusion models contain a \emph{signal leakage}, creating a mismatch between training distribution $q(\bm{x}_{T})$ and inference distribution $q(\hat{\bm{x}}_{T})$. This leakage is pronounced when the diffusion model is tuned on a specific subset of images (middle column), but also exists in the original model (left column). We propose (right column) to realign the inference distribution $q(\hat{\bm{x}}_{T})$ with the training distribution  $q(\bm{x}_{T})$, by modeling the distribution of the signal leak via $\Tilde{q_0} \approx q(\bm{x}_{0})$. (For visualization purposes, the plots consider images as 1D data points, their data distributions $q(\bm{x}_{0})$ are chosen arbitrarily, and the noise schedule used here accentuates the discrepancy.)  }
   \label{fig:teaser}
\end{strip}
\begin{abstract}
    There is a bias in the inference pipeline of most diffusion models. This bias arises from a signal leak whose distribution deviates from the noise distribution, creating a discrepancy between training and inference processes. We demonstrate that this signal-leak bias is particularly significant when models are tuned to a specific style, causing sub-optimal style matching. Recent research tries to avoid the signal leakage during training.
    We instead show how we can exploit this signal-leak bias in existing diffusion models to allow more control over the generated images. This enables us to generate images with more varied brightness, and images that better match a desired style or color.
    By modeling the distribution of the signal leak in the spatial frequency and pixel domains, and including a signal leak in the initial latent, we generate images that better match expected results without any additional training. 
\end{abstract}
\vspace{-0.5cm}
\section{Introduction}

    Denoising diffusion models \cite{ho2020denoising} employ a sequential denoising process to generate visually appealing images from noise. During training, real images are corrupted with white noise, and the diffusion model is tasked to denoise the corrupted images back to their uncorrupted versions. During inference, the trained diffusion model is given white noise, which it progressively denoises to generate realistic images.

    Interestingly, during training, images are corrupted to various degrees, but in the case of most currently available models, if not all, images are never corrupted down to \emph{complete}  noise~\cite{lin2023common, guttenberg2023diffusion}. Even at the last timestep, when the noise level is maximal, corrupted images contain a \emph{signal leak}, \ie they are not composed only of noise but still contain a part of the original real images~\cite{lin2023common}. Only a limited number of studies examine whether the inference process actually matches the training process~\cite{lin2023common, ning2023input, li2023alleviating, salimans2022progressive}, and only a few~\cite{lin2023common, guttenberg2023diffusion, li2023alleviating}~explicitly mention this issue. Following this observation, we argue that starting denoising from only noise at inference time is not aligned with the training process and often results in a \emph{signal-leak bias}.
    
    For instance, images generated with Stable Diffusion~\cite{rombach2022high} tend to always have a medium brightness~\cite{lin2023common, guttenberg2023diffusion}. This bias occurs because the model learns to utilize the brightness of the signal leak to infer the brightness of the real image. At inference time, \emph{starting denoising from white noise is biased} toward generating images with medium brightness, because white noise, which the model interprets as the signal leak, has a medium brightness. Likewise, sampling the initial latent from white noise also biases the generated images to have medium low-frequency components in general, \ie colors and brightness tend to be similar in different areas of the image. More importantly, we notice that the signal-leak bias \emph{prevents models tuned on specific styles from faithfully reproducing the desired styles}.
    
    Aware of the existence of this bias, recent research~\cite{lin2023common, guttenberg2023diffusion} proposes to fine-tune the diffusion models to reduce or remove the signal leak during training, hence generating images with more varied brightness.
    We, on the contrary, propose to \emph{exploit the signal-leak bias to our advantage}.  Instead of fine-tuning or retraining models to eliminate the bias, our approach consists in estimating the distribution of the signal leak with a simple distribution $\Tilde{q_0} \approx q(\bm{x}_0)$. This is done by computing statistics on a small set of target images, for instance, the mean and covariance of their low-frequency content, or the mean and element-wise variance of the pixel values. During inference, rather than denoising from a latent made only of white noise, we \emph{start denoising from a latent composed of both white noise and signal leak}, exactly like during the training of the diffusion model.

    Figure \ref{fig:teaser} visually depicts the source of the signal-leak bias. The middle column shows intuitively the discrepancy between training and inference caused by the signal leakage when the model is tuned for a specific style. Our approach shown in the right column, mitigates this discrepancy, by introducing a signal leak $\sqrt{\bar{\alpha}_{T}} \tilde{\bm{x}}, \Tilde{\bm{x}} \sim \Tilde{q_0}$ in the initial latents $\hat{\bm{x}}_{T}$ during inference, hence mirroring the training distribution. 

    \noindent Our contributions are thus as follows:
    \begin{itemize}[nosep, leftmargin=15pt]
        \item We provide an analysis of the signal-leak bias, with new insights on its origin and its implications. (Section \ref{sec:theory})
        \item We propose a novel approach to include a signal leak in the sampling of the initial latents, instead of sampling them from noise only, biasing the generated images towards generating specific features. (Section \ref{sec:method})
    \end{itemize}
    
    \noindent We show in this paper how to use this approach and leverage the signal-leak bias to our advantage to:
    \begin{itemize}[nosep, leftmargin=15pt]
        \item significantly enhance the outcome of models tuned on images of a specific style, without any additional fine-tuning. (Section \ref{sec:finetuning})
        \item generate images in a particular style with Stable Diffusion \cite{rombach2022high} without any fine-tuning. (Section \ref{sec:notuning})
        \item obtain more diverse images with Stable Diffusion \cite{rombach2022high}, fixing the issue of generated images having a medium brightness, and this without any training. (Section \ref{sec:varied_images})
        \item provide greater control over the generated images, allowing to generate images with specific mean color, without any training. (Section \ref{sec:control})
    \end{itemize}
\section{Background and related work}

\subsection{Denoising Diffusion Probabilistic Models}
    
    Diffusion models \cite{ho2020denoising} learn to denoise corrupted versions $\bm{x}_{t}$ of images $\bm{x}_0 \sim q(\bm{x}_0)$. 
    The noising diffusion process comprises $T$ timesteps, typically $T=1000$. At the first timestep $t=1$, the image $\bm{x}_{1}$ is a slightly noisy version of $\bm{x}_0$. At the last timestep $t=T$, the image $\bm{x}_{T}$ is almost indistinguishable from noise. Transitions from $\bm{x}_0$ to $\bm{x}_{T}$ are parameterized by a noise schedule, \ie a function $\alpha_{t}$ of the timestep $t$.
    For any timestep $t\in[1, T]$, the noisier version $\bm{x}_{t}$ of $\bm{x}_{t-1}$ is obtained from the conditional distribution:
    \begin{gather}
        \nonumber \\[-27pt]
        \label{eq:forward}
        q(\bm{x}_{t} | \bm{x}_{t-1}) = \mathcal{N}(\bm{x}_{t}; \sqrt{\alpha_{t}} \bm{x}_{t-1}, (1-\alpha_{t}) \bm{I}) \\
        \text{\ie~} \bm{x}_{t} = \sqrt{\alpha_{t}} \bm{x}_{t-1} + \sqrt{1-\alpha_{t}} \bm{\varepsilon}, \quad \bm{\varepsilon} \sim  p_{\text{noise}}=\mathcal{N}(\bm{0}, \bm{I})  \nonumber
    \end{gather}
    This describes a first-order Markov chain. Using the notation $\bar{\alpha}_{t} = \prod_{s=1}^t \alpha_{s}$, we have by the chain rule:
    \begin{gather}
        \label{eq:forward_sampling} q(\bm{x}_{t} | \bm{x}_{0}) = \mathcal{N}(\bm{x}_{t}; \sqrt{\bar{\alpha}_{t}} \bm{x}_0, (1- \bar{\alpha}_{t}) \bm{I}) \\
        \text{\ie~} \bm{x}_{t} = \sqrt{\bar{\alpha}_{t}} \bm{x}_0 + \sqrt{1- \bar{\alpha}_{t}} \bm{\varepsilon}, \quad \bm{\varepsilon} \sim  p_{\text{noise}} \nonumber
    \end{gather}
    
    Diffusion models are trained to reverse the forward process described in Equations \ref{eq:forward} and \ref{eq:forward_sampling}. Namely, a neural network $q_{\theta, t}(\bm{x}_{t-1} | \bm{x}_{t})$ with learnable parameters $\theta$ is trained to predict (the distribution of) $\bm{x}_{t-1}$ from a sample $\bm{x}_{t}$. With some reparametrizations, the neural network can be trained to predict (the distribution of) $\bm{\varepsilon}$ knowing $\bm{x}_{t}$, $\bm{x}_{0}$ knowing $\bm{x}_{t}$, or $\sqrt{\bar{\alpha}_{t}} \bm{\varepsilon} - \sqrt{1- \bar{\alpha}_{t}} \bm{x}_{0}$ knowing $\bm{x}_{t}$. These correspond to epsilon-prediction \cite{ho2020denoising}, sample-prediction \cite{salimans2022progressive}, and velocity-prediction \cite{salimans2022progressive}, respectively.    
    
    Assuming epsilon-prediction, one training iteration of the neural network $\bm{\varepsilon}_{\theta, t}(\bm{x}_{t})$ is as follows. An image $\bm{x}_0$ of the dataset, a random timestep $t \sim \mathcal{U}([1, T])$ and a noise $\bm\varepsilon \sim p_{\text{noise}}$ are sampled. A noise-contaminated image $\bm{x}_{t}$ is built according to Equation \ref{eq:forward_sampling}.
    The neural network is given $\bm{x}_{t}$ and $t$, and outputs a predicted noise $\bm{\varepsilon}_{\theta, t}(\bm{x}_{t})$. The loss for epsilon-prediction is typically set as the mean square error $|| \bm{\varepsilon}_{\theta, t}(\bm{x}_{t}) - \bm{\varepsilon} ||_2^2$.
    
    At inference time, an initial latent $\hat{\bm{x}}_{T}$ is sampled from: \\
    \begin{gather}
        \nonumber \\[-28pt]
        \label{eq:initial_latent}
        q(\hat{\bm{x}}_{T}) = p_{\text{noise}}=\mathcal{N}(\bm{0}, \bm{I}) \\
        \text{\ie~} \hat{\bm{x}}_{T} = \bm\varepsilon, \quad \bm\varepsilon \sim p_{\text{noise}} \nonumber
    \end{gather}
    and iteratively denoised. For any timestep $t\in[1, T]$:
    \begin{gather}
        \label{eq:denoising}
        q_\theta(\hat{\bm{x}}_{t-1}) = \textstyle{\int} q(\hat{\bm{x}}_{T}) \textstyle{\prod}_{s=t}^{T} q_{\theta, s}(\hat{\bm{x}}_{s-1} | \hat{\bm{x}}_{s}) d\hat{\bm{x}}_{t:T} \\
         \text{thus, to sample each } \hat{\bm{x}}_{t-1} \text{, we use } \hat{\bm{x}}_{t} \text{ and } q_{\theta, t}(\hat{\bm{x}}_{t-1} | \hat{\bm{x}}_{t})  \nonumber
    \end{gather}
    
    Instead of denoising through all the timesteps of the model to generate an image $\hat{\bm{x}}_{0}$, accelerated sampling algorithms have been proposed \cite{song2020denoising, liu2022pseudo}, reducing the number of forward passes through the neural network $q_{\theta, t}$ by orders of magnitude, \eg from 1000 to 50 \cite{liu2022pseudo}. In such cases, the first denoising iteration does not always start denoising from the highest timestep $t = 1000$, but for instance $t = 981$ (see details in Section 3.3. of \cite{lin2023common}).
    Without loss of generality, we can assume in such a case that the model was only trained with $T=981$ timesteps and that the inference process always starts denoising from the last timestep $T$.
    
    A commonly used diffusion model is Stable Diffusion, which generates high-quality images conditionally on a textual prompt. Stable Diffusion is a Latent Diffusion Model (LDM, \cite{rombach2022high}), meaning the images are represented in a latent space instead of the pixel space. The equations above still hold if we consider $\bm{x}_0$ to be an image represented by a latent code. In particular, latent codes in LDMs still have channels and two spatial dimensions. We thus keep the terminology \emph{pixel} to refer to an element of the latent code.

\subsection{Fixing the training of diffusion models}
\label{sec:related_fixing}

    Sampling the initial latents from noise only (Equation \ref{eq:initial_latent}) is not totally aligned with the training process (Equation \ref{eq:forward_sampling}), where images are not corrupted up to complete noise \cite{guttenberg2023diffusion}, but always contain a signal leak $\sqrt{\bar{\alpha}_{T}} \bm{x}_0$ from a real image $\bm{x}_0$ \cite{lin2023common}.
    In the case of Stable Diffusion, for which the signal leakage is particularly important \cite{lin2023common}, this leads for instance to generated images with medium brightness \cite{lin2023common, guttenberg2023diffusion}.
    To eliminate this issue, recent research \cite{salimans2022progressive, lin2023common} trains diffusion models enforcing $\alpha_T=0$, effectively training the last timestep from white noise, \ie without signal leakage. Guttenberg \cite{guttenberg2023diffusion} proposes to modify the noise distribution, 
    such that the 
    brightness of the real image cannot be deduced from the signal leak anymore.
    The signal leakage also leads to difficulties in generating style-specific images. To overcome this, Everaert \etal \cite{diffusion_in_style} propose to finetune Stable Diffusion on a new noise distribution that approximates the distribution of the style images.
    While not focusing on the signal leakage, Ning \etal \cite{ning2023input} propose adding an extra perturbation term during training to make the model more robust to training and inference distribution changes. All of these require retraining/finetuning the models to remove or reduce any signal leakage.
    
    Our approach, on the contrary, can be used directly with any existing diffusion model. It leverages the signal leakage instead of retraining or fine-tuning models. Rather than finetuning to realign the training distribution with the inference distribution, \ie the distribution of initial latents, we focus on realigning the inference distribution with the training distribution, by adding a signal leak $\sqrt{\bar{\alpha}_{T}} \tilde{\bm{x}}$ to the noise $\bm{\varepsilon}$ in the initial latents $\hat{\bm{x}}_T$. 
\section{Signal-leak bias}
\label{sec:theory}

    \subsection{Discrepancies between training and inference distributions in diffusion models}
    
    The reverse diffusion (\ie denoising) process described in Equation \ref{eq:denoising} inputs the neural network $q_{\theta, t}$ with some data $\hat{\bm{x}}_{t}$ to obtain $\hat{\bm{x}}_{t-1}$. However, $\hat{\bm{x}}_{t}$ is obtained either from previous predictions of the model when $t<T$ or, if $t=T$, from white noise. This differs from training, where $\bm{x}_{t}$ is a corrupted version of a real image $\bm{x}_{0}$. In both cases, $t<T$ and $t=T$, the inference distribution differs from the training distribution. 

    \paragraph{Exposure bias:} The diffusion model $q_{\theta, t}$ is trained using noise-corrupted versions of images (Equation \ref{eq:forward_sampling}), rather than with the predictions of the latter timesteps as done during inference (Equation \ref{eq:denoising}). This creates a discrepancy between training and inference, which can cause error accumulation during the iterative denoising process to generate images \cite{ning2023input, li2023alleviating}, similarly to the \emph{exposure bias} \cite{ranzato2016sequence, schmidt2019generalization} in text-generation models.    
    \vspace{3pt}
    
    Due to the signal leakage, a significant ``error'' often already exists at the last timestep ($T$), as we explained below. This error will be accumulated forward through the image generation process, and hence should not be ignored. 

    \paragraph{Signal-leak bias:} At inference time, the model is given white noise as initial latent $\hat{\bm{x}}_{T}$ (Equation \ref{eq:initial_latent}). However, we can deduce from Equation \ref{eq:forward_sampling} that the model was trained at the last timestep with samples $\bm{x}_{T}$ from $q(\bm{x}_{T})$:
    \begin{align}
        \label{eq:training_last}    
        \bm{x}_{T} \sim q(\bm{x}_{T}) \Leftrightarrow &~ \bm{x}_{T} = \sqrt{\bar{\alpha}_{T}} \bm{x}_0 + \sqrt{1- \bar{\alpha}_{T}} \bm\varepsilon, \\ 
        &\bm{x}_0 \sim q(\bm{x}_0), \quad \bm\varepsilon \sim p_{\text{noise}} \nonumber
    \end{align}
    The two distributions $q(\hat{\bm{x}}_{T})$ and $q(\bm{x}_{T})$ differ, creating a discrepancy between training and inference distributions. Following Lin \etal \cite{lin2023common}, we can quantify the importance of the signal leakage by a signal-to-noise ratio (SNR):
    \begin{gather}
        \label{eq:snr}    
        \text{SNR} = \bar{\alpha}_{T} / (1-\bar{\alpha}_{T})
    \end{gather}
    This SNR depends on $\bar{\alpha}_{T}$, and hence on the choice of the function $\alpha_t$ \cite{lin2023common}. The function $\alpha_t$ is defined according to a $\beta$-schedule. The function $\beta_t = 1-\alpha_t$ is typically chosen to be 
    a linear schedule in $\beta$-space \cite{ho2020denoising},
    a squared capped cosine schedule in $\beta$-space \cite{nichol2022glide, nichol2021improved}, 
    a linear schedule in $\sqrt\beta$-space \cite{rombach2022high}, 
    or a sigmoid schedule in $\beta$-space \cite{xugeodiff}. 
    The SNR is particularly high with Stable Diffusion \cite{lin2023common}, which uses the linear schedule in $\sqrt\beta$-space. Lin \etal \cite{lin2023common} also link the signal leakage to the fact that Stable Diffusion always generates images with medium brightness.    
    
    We notice that this 
    linear schedule in $\sqrt\beta$-space is not only used in Stable Diffusion, but is commonly employed in many variants of LDM \cite{rombach2022high} as well, \eg \cite{stan2023ldm3d, chang2023flex, huggingfaceCerspensezeroscopev2XLHugging}.

\subsection{Mismatch between noise and signal leak distributions}
    \begin{figure}[b]
       \centering
       \vspace{-3pt}
       \setlength{\tabcolsep}{.5pt}
        \begin{tabular}{ccccc}
    
            \makecell[c]{} &
            \multicolumn{3}{c|}{Training}& 
            \makecell[c]{Inference}\\
            &
            \makecell[c]{$\bm{x}_0$} &
            \makecell[c]{$\bm{x}_{500}$} &
            \multicolumn{1}{c|}{\makecell[c]{$\bm{x}_{1000}$}} &
            \makecell[c]{$\hat{\bm{x}}_{1000}$} \vspace{1pt} \\

            \makecell[c]{$\bm{x}$} &
            \includegraphics[align=c, width=.18\linewidth]{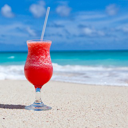} & 
            \includegraphics[align=c, width=.18\linewidth]{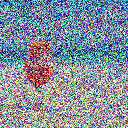} & 
            \includegraphics[align=c, width=.18\linewidth]{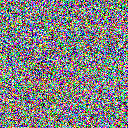} & 
            \includegraphics[align=c, width=.18\linewidth]{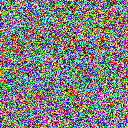}\vspace{1pt} \\
            
            \makecell[c]{$\bm{X}\mkern-2mu{=}\mkern1mu\text{DCT}(\bm{x})$ \\ \footnotesize (log-scale, \\[-3pt] \footnotesize $u \text{ and } v < 32$)} &
            \includegraphics[align=c, width=.18\linewidth]{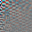} & 
            \includegraphics[align=c, width=.18\linewidth]{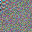} & 
            \includegraphics[align=c, width=.18\linewidth]{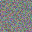} & 
            \includegraphics[align=c, width=.18\linewidth]{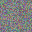}\vspace{1pt}\\
            
            \makecell[c]{IDCT of \\ $M_1 \odot \frac{\bm{X}}{\sqrt{\bar{\alpha}_{t}}}$} &
            \includegraphics[align=c, width=.18\linewidth]{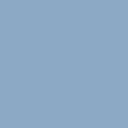} & 
            \includegraphics[align=c, width=.18\linewidth]{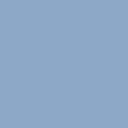} & 
            \includegraphics[align=c, width=.18\linewidth]{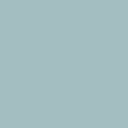} & 
            \includegraphics[align=c, width=.18\linewidth]{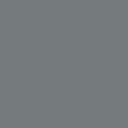}\vspace{1pt}\\
            
            \makecell[c]{IDCT of \\ $M_3 \odot \frac{\bm{X}}{\sqrt{\bar{\alpha}_{t}}}$} &
            \includegraphics[align=c, width=.18\linewidth]{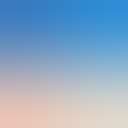} & 
            \includegraphics[align=c, width=.18\linewidth]{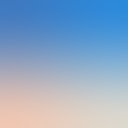} & 
            \includegraphics[align=c, width=.18\linewidth]{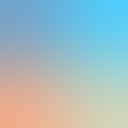} & 
            \includegraphics[align=c, width=.18\linewidth]{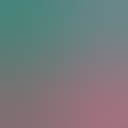}\\
            
            \makecell[c]{$t$} &
            \makecell[c]{$0$} & 
            \makecell[c]{$500$} & 
            \makecell[c]{$1000$} & 
            \makecell[c]{$1000$}\\

        \end{tabular}
       \vspace{-4pt}
       \caption{
       The first row shows values for  $\bm{x}_0$,  $\bm{x}_{500}$,  $\bm{x}_{1000}$, and $\hat{\bm{x}}_{1000}$. The second row contains their 2D-DCT components, showing that natural images mostly contain low-frequency components, unlike white noise, which is equally spread across all frequencies. 
       As shown in the third and fourth rows, we can recover some low-frequencies components of the original image $\bm{x}_0$ from $\bm{x}_{500}$ or partially from $\bm{x}_{1000}$. By eliminating all frequencies except the lowest one (third row) or the 3 lowest ones (fourth row), we successfully recover the low-frequency information of $\bm{x}_0$ shown in the first column, third and fourth rows. The noise introduced during the diffusion process does not affect these low-frequency components; thus allowing the model to learn not to alter them during denoising, with the result of generating images with similar low-frequency content as white noise, \eg medium brightness. $M_N$ refers to the mask of the $N$ lowest-frequencies and $\odot$ symbolizes element-wise multiplication. IDCT refers to the Inverse Discrete Cosine Transform. We used the same $\beta$-schedule as LDM \cite{rombach2022high}.   }
       \label{fig:dct}
    \end{figure}

    \paragraph{The source of the signal-leak bias:}  
    
    The strength of the signal leakage depends on $\bar{\alpha}_{T}$. If $\bar{\alpha}_{T}=0$, then there is no signal leakage at the last timestep $T$, as can be deduced from Equation \ref{eq:forward_sampling}. This is rarely the case \cite{salimans2022progressive, lin2023common}.
    
    When $\bar{\alpha}_{T} \neq 0$, the signal leakage exists, but does not necessarily imply a bias at inference time. 
    As seen by the high quality of the images generated by LDMs~\cite{rombach2022high, stan2023ldm3d, chang2023flex, huggingfaceCerspensezeroscopev2XLHugging}, the existence of the signal leak can have little implication in practice. In LDMs, $\bar{\alpha}_{T}$ is high, leading to $\bm{x}_{T} = 0.068265 \cdot \bm{x}_0 + 0.997667 \cdot \bm\varepsilon$ \cite{lin2023common}.
    Even though $\bar{\alpha}_{T}$ is high, the training distribution $q(\bm{x}_{T})$ and inference distribution $q(\hat{\bm{x}}_{T})$ are relatively aligned because the noise distribution $p_{\text{noise}}$ and the image distribution $q(\bm{x}_0)$ are similar. Indeed, in LDMs \cite{rombach2022high}, the diffusion happens in a normalized VAE latent space \cite{kingma2013auto}, where images $\bm{x}_0$ are represented by their VAE latent codes. The VAE makes the distribution of images $q(\bm{x}_0)$ relatively similar to the noise distribution $p_{\text{noise}} = \mathcal{N}(\bm{0}, \bm{I})$. The signal leak $\sqrt{\bar{\alpha}_{T}} \bm{x}_0$ is then almost indistinguishable from noise, leading to almost no bias when the initial latents are sampled from noise only.     
    
    However, whenever the signal leak has a distribution that differs from the noise distribution, sampling the initial latent from only noise creates a bias. At inference time, the model expects to find a signal leak $\sqrt{\bar{\alpha}_{T}} \bm{x}_0$ in the initial latent $\hat{\bm{x}}_T$ to deduce information about the real image $\bm{x}_0$. Sampling the initial latents from noise only biases the generated images, because the model interprets noise as being the signal leak. This is especially noticeable when trying to generate images in a specific style: the model expects to find an initial latent from a distribution that is different from white noise (see the middle column of Figure \ref{fig:teaser}).

    \paragraph{Why Stable Diffusion always generates images with medium brightness:}  
    In Stable Diffusion, the distribution of images $q(\bm{x}_0)$ does not \emph{exactly} match $p_{\text{noise}} = \mathcal{N}(\bm{0}, \bm{I})$. For example, the mean of the pixels of a sample from $p_{\text{noise}}$ always has a medium value, but the mean of pixels of a real image will be more varied, depending on the brightness of the image. This causes images generated with Stable Diffusion to always have a medium brightness \cite{lin2023common, guttenberg2023diffusion}. We discuss here the mismatch between $q(\bm{x}_0)$ and $p_{\text{noise}}$ from a \emph{frequency domain} point of view.
    Natural images tend to be smooth and exhibit an average power spectrum that declines with a $f^{-2}$ relationship \cite{tolhurst1992amplitude, ruderman1994statistics, burton1987color, field1987relations}, indicating a concentration of signal power at the lowest spatial frequencies.
    On the other hand, the white noise $\mathcal{N}(\bm{0}, \bm{I})$ is equally distributed across all frequencies, meaning noising mostly affects high-frequency components \cite{el2020stochastic}. Rewriting the Equation \ref{eq:snr} for a specific spatial frequency $(u,v)$, we obtain: 
    \begin{align}
        \label{eq:snr_freq}    
        \text{SNR}(u,v) &= \frac{\bar{\alpha}_{T} \mathbb{E}_{\bm{x}_0 \sim q(\bm{x}_0)}(({\bm{X}_0^{u,v}})^2)} {(1-\bar{\alpha}_{T})\mathbb{E}_{\bm\varepsilon \sim \mathcal{N}(\bm{0}, \bm{I})}((\bm{\mathcal{E}}^{u,v})^2)} \\
        &= \frac{\bar{\alpha}_{T}} {1-\bar{\alpha}_{T}} \mathbb{E}_{\bm{x}_0 \sim q(\bm{x}_0)}(({\bm{X}_0^{u,v}})^2) \nonumber
    \end{align}
    where $\bm{X}_0^{u,v}$ denotes the $(u, v)$-th term of the 2D-DCT of an image $\bm{x}_0$. Note that $\bm{\mathcal{E}}^{u,v}$, the $(u, v)$-th term of the 2D-DCT of a noise sample $\bm\varepsilon \sim \mathcal{N}(\bm{0}, \bm{I})$, also follows $\mathcal{N}(\bm{0}, \bm{I})$. 
    Because of the prevalence of low spatial frequency content, $\mathbb{E}_{\bm{x}_0}(({\bm{X}_0^{u,v}})^2)$ is high when $u$ and $v$ are small, and negligible for high frequencies. The SNR is then high for the lowest frequencies and almost 0 for the remaining frequencies. These observations can be visualized as in Figure~\ref{fig:dct}.        
    Note that the mean color of the image $\bm{x}_0$, \ie the signal $\bm{X}_0^{0,0}$, is thus the least affected by the noise. The diffusion model then learns to recover the mean color $\bm{X}_0^{0,0}$ of the image $\bm{x}_0$ from the one $\bm{X}_T^{0,0}$ of $\bm{x}_T$. When $\hat{\bm{x}}_T$ is sampled from $\mathcal{N}(\bm{0}, \bm{I})$, then $\hat{\bm{X}}_T^{0,0} \approx 0$, thus generated images $\hat{\bm{x}}_0$ always result in $\bm{\hat{X}}_0^{0,0} \approx 0$, \ie, a medium brightness.    

    \paragraph{Limitation of Stable Diffusion after tuning on a style:}  
    Fine-tuning Stable Diffusion to a specific style usually does not work as intended. Generated images do not match the colors or backgrounds of the style, as illustrated in the first rows of Figures \ref{fig:results_pokemon}, \ref{fig:results_lineart}, and \ref{fig:results_nasa}. Even when fine-tuned on a single solid black image, Stable Diffusion is unable to produce a black image \cite{guttenberg2023diffusion}. Everaert \etal \cite{diffusion_in_style} show that training with style-specific noise instead of white noise leads to better style adaptation. 
    We then argue that, when fine-tuning for a specific style \textit{with white noise}, there is a significant mismatch between noise and image distributions \emph{in the pixel domain}. 
    The distribution of images of a specific style is located in a specific part of the image space. Hence $q(\bm{x}_0)$ cannot be considered similar to $\mathcal{N}(\bm{0}, \bm{I})$ anymore. Because of the signal leakage, the training distribution $q(\bm{x}_{T})$ is far from the inference distribution $q(\hat{\bm{x}}_{T})$. Images generated by a diffusion model tuned to a specific style thus do not look as good as they potentially could.

\section{Method}
\label{sec:method}

\subsection{Exploiting the signal-leak bias}

    As discussed in Section \ref{sec:related_fixing}, previously proposed solutions mainly focus on eliminating the signal-leak bias by setting $\alpha_T=0$ \cite{salimans2022progressive, lin2023common}, on adding noise perturbations \cite{ning2023input}, or on modifying noise distribution \cite{guttenberg2023diffusion, diffusion_in_style}. Essentially, these methods attempt to realign the training distribution with the inference distribution. This comes at the cost of re-training or fine-tuning a model. To our knowledge, only Li \etal \cite{li2023alleviating} propose a solution to realign the distributions at inference time, without re-training. At each denoising iteration, they propose to find the best timestep $t'$ at which to denoise the current $\hat{\bm{x}}_{t}$. However, while efficient for the exposure bias, this cannot work for the signal-leak bias - there is simply no timestep $t'$ trained with samples from $\mathcal{N}(\bm{0}, \bm{I})$.
    
    Our solution to \textit{exploit the signal-leak bias in diffusion models} is much simpler than these previous solutions. We focus on \emph{realigning the distribution of initial latent $q(\hat{\bm{x}}_{T})$ with the training distribution $q({\bm{x}}_{T})$}. This has the advantage of not requiring any additional training of the diffusion model.    
    The key idea of our solution is to simply sample the initial latents $\hat{\bm{x}}_{T}$ from the training distribution (Equation \ref{eq:training_last}) instead of from only white noise. Although the distribution $q(\bm{x}_0)$ is unknown, we can approximate it by computing statistics from a set of target images. 
    Our approach then consists of obtaining an approximate distribution  $\Tilde{q_0}$ of $q(\bm{x}_0)$. At inference time, we simply sample the initial latents in the same way as during training (Equation~\ref{eq:forward_sampling}), \ie with a random signal leak $\sqrt{\bar{\alpha}_{T}} \Tilde{\bm{x}}$: 
    \begin{gather}
        \label{eq:new_dist}    
        \hat{\bm{x}}_{T} = \sqrt{\bar{\alpha}_{T}} \Tilde{\bm{x}} + \sqrt{1- \bar{\alpha}_{T}} \bm\varepsilon \\
        \Tilde{\bm{x}} \sim \Tilde{q_0}, \quad \bm\varepsilon \sim p_{\text{noise}}=\mathcal{N}(\bm{0}, \bm{I}) \nonumber 
    \end{gather} 
    
    Note that no other operations are needed and we then just follow the usual process of generating images with diffusion models. Equation \ref{eq:new_dist} has similarities with the image-editing work SDEdit \cite{meng2021sdedit}, which samples intermediate latents $\hat{\bm{x}}_{t_0}$ as $\sqrt{\bar{\alpha}_{t_0}} \bm{x}^{(g)} + \sqrt{1- \bar{\alpha}_{t_0}} \bm\varepsilon, \bm\varepsilon \sim p_\text{noise}$ (\ie Equation \ref{eq:forward_sampling}), where $\bm{x}^{(g)}$ is an image to be edited. SDEdit \cite{meng2021sdedit} focuses only on image editing and uses $t_0 \approx 0.3T\text{ to }0.6T$. Note that in our work, unlike SDEdit, we generate images starting from the timestep $T$. 
    
\subsection{Modeling the distribution of the signal leak}
\label{sec:method_choice}
    The current image generation process of diffusion models, which samples $\hat{\bm{x}}_{T}$ from $\mathcal{N}(\bm{0}, \bm{I})$, is equivalent to using Equation \ref{eq:new_dist} with $\Tilde{q_0} = \mathcal{N}(\bm{0}, \bm{I})$. We now discuss better choices for $\Tilde{q_0}$. Following our previous insights, we conclude that the signal leak mismatches the noise distribution either in the \emph{pixel domain} or in the \emph{frequency domain}.
    
    We provide here two models for the distribution of the signal leak. The first one estimates the distribution of the signal leak in the pixel domain. We use it in Sections \ref{sec:finetuning} and \ref{sec:notuning}. The second one, used in Sections \ref{sec:varied_images} and \ref{sec:control}, estimates the distribution of the signal leak in the frequency domain \emph{and} in the pixel domain, for the low-frequency (LF) \emph{and} high-frequency (HF) contents, respectively.    
    To be more specific, we provide the dimensions of the elements for Stable Diffusion V1, where latent codes of images have $4$ channels and $64 \times 64$ pixels, \ie $\bm{x}_0 \in \mathbb{R}^{64\times64\times4}$. These values are to be adapted to the model being used and do not imply that our approach requires specific architectural changes.

    \subsubsection{Pixel-domain model}
    \label{sec:pixel_domain}
    We first model the approximate distribution $\Tilde{q_0}$ as $\mathcal{N}(\bm{\mu}, \text{diag}(\bm{\sigma}^2))$, a Gaussian distribution with diagonal covariance. The location $\bm{\mu}$ and covariance $\text{diag}(\bm{\sigma}^2)$ are obtained from statistics of the target images.
    \begin{align}
        (\bm{\mu})^{i, j, k} &= \text{Mean}_{\bm{x}_0}~\bm{x}_0^{i, j, k} \\
        (\bm{\sigma})^{i, j, k} &= \text{Std}_{\bm{x}_0}~\bm{x}_0^{i, j, k} \\
        \omit\rlap{$\bm{\mu} \in \mathbb{R}^{64\times64\times4},~ \bm{\sigma} \in \mathbb{R}^{64\times64\times4}$} \nonumber 
    \end{align}

    These equations have similarities with prior research~\cite{diffusion_in_style, zhou2023shifted}. However, we only use this distribution to model the signal leak $\sqrt{\bar{\alpha}_{T}} \bm{x}_0$, instead of training the model with it.    
    As we show in Sections \ref{sec:finetuning} and \ref{sec:notuning}, this \emph{pixel-domain model} of the signal leak is effective for style adaptation of diffusion models. Yet, because the distribution of natural images in LDMs \cite{rombach2022high} is already approximately $\mathcal{N}(\bm{0}, \bm{I})$, this pixel-domain model does not help to generate images with more varied brightness.
    
    \subsubsection{Frequency and pixel domain model}
    \label{sec:freq_domain}
    As mentioned before, the training distribution of diffusion models on natural images differs from the inference distribution mostly in the lowest frequencies. We can thus explicitly model the $N$ lowest frequencies of the signal leak by computing the mean and covariance of the low-frequency components from a small set of natural images. The remaining, \ie the components with higher frequencies, are modeled in the pixel domain as in the previous paragraph.
    To model the $N$ lowest frequencies, we compute the DCT $\bm{X}_{0}$ of each target image $\bm{x}_{0}$. We obtain a multivariate Gaussian distribution $\Tilde{q_{0}}_\text{LF}$ with location $\bm{\mu}_\text{LF}$ and covariance $\bm{\Sigma}_\text{LF}$ by computing statistics from the DCTs $\bm{X}_{0}$. 
    With the notation $\bm{X}_{0, \text{LF}}=M_N \odot \bm{X}_{0}$, the location and covariance are estimated as follows:
    \begin{align}
        \Tilde{q_{0}}_\text{LF} &= \mathcal{N}(\bm{\mu}_\text{LF}, \bm{\Sigma}_\text{LF}) \\
        (\bm{\mu}_\text{LF})^{u, v, k} &= \text{Mean}_{\bm{x}_0}~\bm{X}_{0, \text{LF}}^{u, v, k} \nonumber \\
        (\bm{\Sigma}_\text{LF})^{u_1, v_1, k_1, u_2, v_2, k_2} &= \text{Cov}_{\bm{x}_0}(\bm{X}_{0, \text{LF}}^{u_1, v_1, k_1}, \bm{X}_{0,\text{LF}}^{u_2, v_2, k_2}) \nonumber \\
        \bm{\mu}_\text{LF} \in \mathbb{R}^{4N}&, \quad \bm{\Sigma}_\text{LF} \in \mathbb{R}^{4N \times 4N} \nonumber 
    \end{align}
    The high-frequency components $\bm{x}_{0,\text{HF}} = \text{IDCT}(\bm{X}_0 - \bm{X}_{0, \text{LF}})$ are modeled with a pixel-domain distribution $\Tilde{q_{0}}_\text{HF}$ as in Section \ref{sec:pixel_domain}. 
    The signal leak $\sqrt{\bar{\alpha}_{T}} \Tilde{\bm{x}}$ that we add in the initial latent at inference time is sampled such that:
    \begin{align}
    \label{eq:fred_new_dist}
        \Tilde{\bm{x}} \sim \Tilde{q_0} \Leftrightarrow ~ &\Tilde{\bm{x}}= \text{IDCT}(\Tilde{\bm{X}}_\text{LF})+\Tilde{\bm{x}}_\text{HF},\\ &\text{with } \Tilde{\bm{X}}_{\text{LF}}\sim \Tilde{q_{0}}_\text{LF} \text{ and } \Tilde{\bm{x}}_\text{HF} \sim \Tilde{q_{0}}_\text{HF} \nonumber
    \end{align}

    By combining the two components LF and HF, we create a distribution $\Tilde{q_0}$ that encompasses a broader range of colour and brightness variations than $\mathcal{N}(\bm{0}, \bm{I})$. Sampling initial latents as in Equation \ref{eq:new_dist} with this estimation $\Tilde{q_0}$ enables the generation of images with more diverse brightness and colours than with  $\mathcal{N}(\bm{0}, \bm{I})$, as we show in Section~\ref{sec:varied_images}.
    The value of $N$ is chosen empirically, for instance, $N=3$.
    While we present results obtained with DCT, note that different approaches to model frequency components could also be used, such as PCA or Fourier Transform.

\section{Results}

We experiment with Stable Diffusion, which has a significant signal leakage \cite{lin2023common}. Following current evaluations of diffusion models \cite{huggingfaceEvaluatingDiffusion, saharia2022photorealistic, rombach2022high},
we compute FID (lower is better, \cite{heusel2017gans}) and CLIP (higher is better, \cite{radford2021learning}) scores from the TorchMetrics library \cite{torchmetrics}. Metrics are computed using 200 generated images. Wherever a CLIP score is reported (Figures \ref{fig:results_pokemon}, \ref{fig:results_lineart}, \ref{fig:results_pokemonnotuning}, \ref{fig:results_lineartnotuning}, and \ref{fig:results_varied}), the 200 images are generated from the textual prompts of the DrawBench benchmark \cite{saharia2022photorealistic}, with a guidance scale of 7.5 \cite{ho2022classifier}. In the other cases (Figures \ref{fig:results_nasa}, \ref{fig:results_nasanotuning}), the 200 images are generated without classifier-free guidance \cite{ho2022classifier}. We compute two versions of the FID, $\text{FID}_{64}$ and $\text{FID}_{2048}$, using the 64-th or 2048-th InceptionV3 \cite{szegedy2016rethinking} feature layers, respectively. All images are generated with 50 PNDM denoising steps \cite{liu2022pseudo}.

\subsection{Improved style for style-specific models}
\label{sec:finetuning}

    \begin{figure}[b]    
    
        \begin{subfigure}[ht]{1.\linewidth}
           \centering
           \setlength{\tabcolsep}{.5pt}
            \begin{tabular}{ccccc}
            
                \makecell[c]{Original \vspace{-1mm} \\ \footnotesize $\text{FID}_{64} = 26.9$ \vspace{-1mm}\\ \footnotesize $\text{FID}_{2048} = 156$ \vspace{-1mm}\\ \footnotesize $\mathbf{\textbf{CLIP} = 29.4}$} &
                \includegraphics[align=c, width=.178\linewidth]{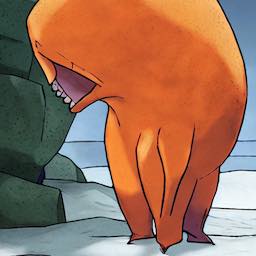} & 
                \includegraphics[align=c, width=.178\linewidth]{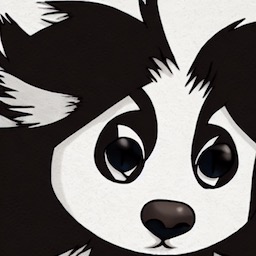} & 
                \includegraphics[align=c, width=.178\linewidth]{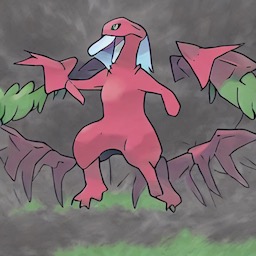} & 
                \includegraphics[align=c, width=.178\linewidth]{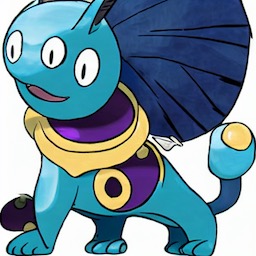} \vspace{1pt} \\
                
                \makecell[c]{Ours \vspace{-1mm} \\\footnotesize $\mathbf{\textbf{FID}_{64} = 1.2}$ \vspace{-1mm}\\ \footnotesize $\mathbf{\textbf{FID}_{2048} = 120}$ \vspace{-1mm}\\ \footnotesize $\text{CLIP} = 27.4$} &
                \includegraphics[align=c, width=.178\linewidth]{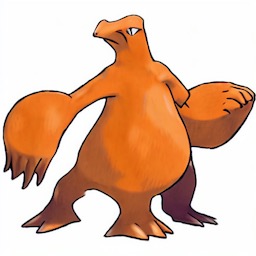} & 
                \includegraphics[align=c, width=.178\linewidth]{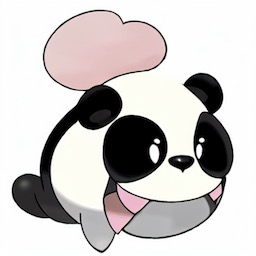} & 
                \includegraphics[align=c, width=.178\linewidth]{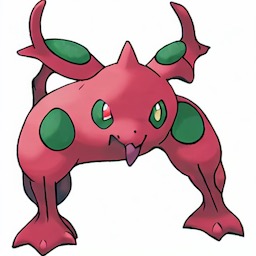} & 
                \includegraphics[align=c, width=.178\linewidth]{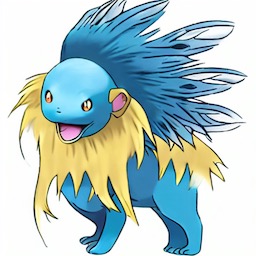} \vspace{1pt}\\
                
            \end{tabular}
           \caption{Pokemon-LoRA model \cite{pokemonLORA} fine-tuned from Stable Diffusion v1.4 on the Pokemon BLIP captions dataset \cite{pokemonblip} with LoRA fine-tuning \cite{hu2021lora, ryu2023low}.}
           \label{fig:results_pokemon}
        \end{subfigure}   
    
        \begin{subfigure}[ht]{1.\linewidth}
           \centering
           \setlength{\tabcolsep}{.5pt}
            \begin{tabular}{ccccc}
        
                \makecell[c]{Original \vspace{-1mm} \\ \footnotesize $\text{FID}_{64} = 35.0$ \vspace{-1mm}\\ \footnotesize $\text{FID}_{2048} = 392$ \vspace{-1mm}\\ \footnotesize $\mathbf{\textbf{CLIP} = 27.5}$} &
                \includegraphics[align=c, width=.178\linewidth]{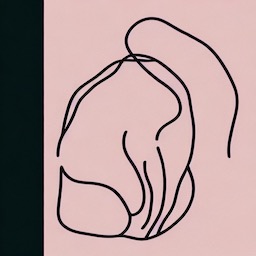} & 
                \includegraphics[align=c, width=.178\linewidth]{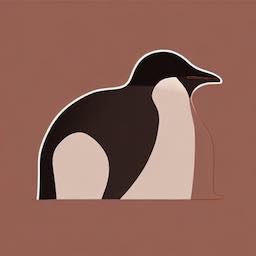} & 
                \includegraphics[align=c, width=.178\linewidth]{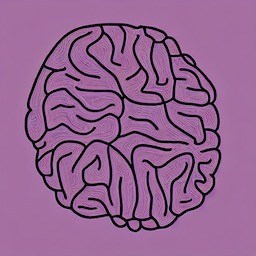} & 
                \includegraphics[align=c, width=.178\linewidth]{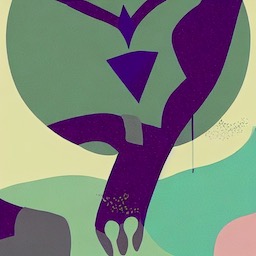} \vspace{1pt} \\
                
                \makecell[c]{Ours \vspace{-1mm} \\\footnotesize $\mathbf{\textbf{FID}_{64} = 2.7}$ \vspace{-1mm}\\ \footnotesize $\mathbf{\textbf{FID}_{2048} = 371}$ \vspace{-1mm}\\ \footnotesize $\text{CLIP} = 27.1$} &
                \includegraphics[align=c, width=.178\linewidth]{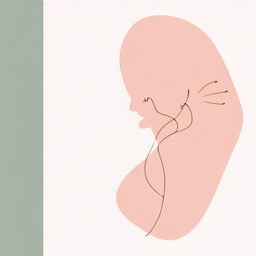} & 
                \includegraphics[align=c, width=.178\linewidth]{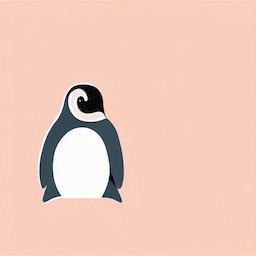} & 
                \includegraphics[align=c, width=.178\linewidth]{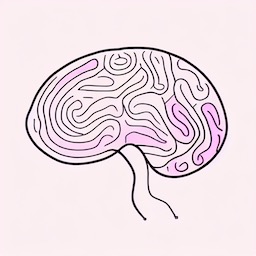} & 
                \includegraphics[align=c, width=.178\linewidth]{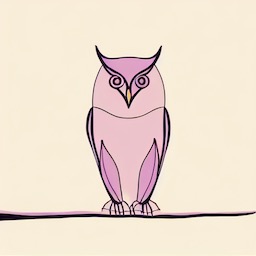} \vspace{1pt}\\
                
            \end{tabular}
           \caption{Stable Diffusion v1.4 model. The concept of ``line-art style'' \cite{lineart} was learned with Textual Inversion \cite{gal2022image}, from  7 line-art images \cite{lineart}.}
           \label{fig:results_lineart}
        \end{subfigure}
    
        \begin{subfigure}[ht]{1.\linewidth}
           \centering
           \setlength{\tabcolsep}{.5pt}
            \begin{tabular}{ccccc}
        
                \makecell[c]{Original \vspace{-1mm} \\ \footnotesize $\text{FID}_{64} = 8.4$ \vspace{-1mm} \\ \footnotesize $\text{FID}_{2048} = 274$} &
                \includegraphics[align=c, width=.178\linewidth]{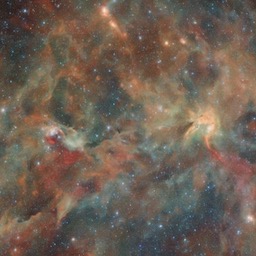} & 
                \includegraphics[align=c, width=.178\linewidth]{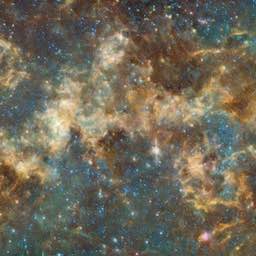} & 
                \includegraphics[align=c, width=.178\linewidth]{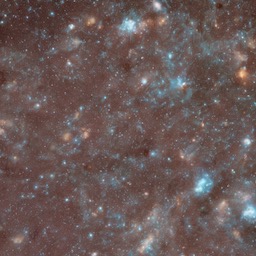} & 
                \includegraphics[align=c, width=.178\linewidth]{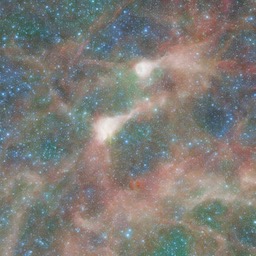} \vspace{1pt} \\
                
                \makecell[c]{Ours \vspace{-1mm} \\ \footnotesize $\mathbf{\textbf{FID}_{64} = 3.8}$ \vspace{-1mm} \\ \footnotesize $\mathbf{\textbf{FID}_{2048} = 156}$} &
                \includegraphics[align=c, width=.178\linewidth]{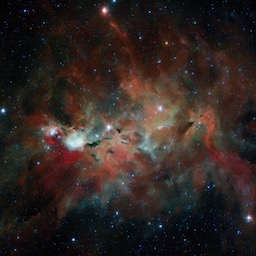} & 
                \includegraphics[align=c, width=.178\linewidth]{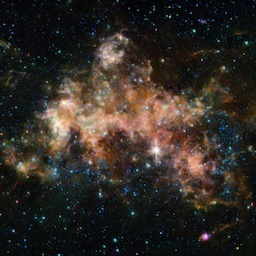} & 
                \includegraphics[align=c, width=.178\linewidth]{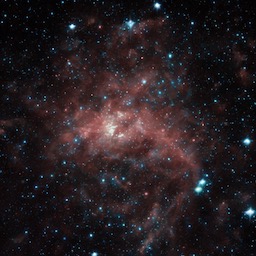} & 
                \includegraphics[align=c, width=.178\linewidth]{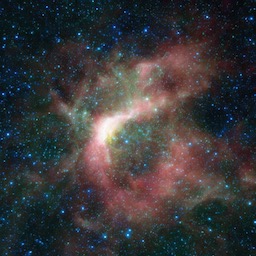} \vspace{1pt}\\
                
            \end{tabular}
           \caption{NASA-space model \cite{nasa} fine-tuned with DreamBooth \cite{ruiz2023dreambooth} from Stable Diffusion v2 on 24 images of astronomical phenomena \cite{nasa}.}
           \label{fig:results_nasa}
        \end{subfigure}
       \caption{ When using white noise as initial latent, existing fine-tuning strategies lead to sub-optimal style-matching, as illustrated in the first row of each subfigure \ref{fig:results_pokemon}, \ref{fig:results_lineart} and \ref{fig:results_nasa}. In each of the second rows, images are generated with the same prompts and the same models, which we did not additionally fine-tune. In this second row, images are generated with our proposed approach, sampling a signal leak $\sqrt{\bar{\alpha}_{T}} \tilde{\bm{x}}$ from our pixel-domain estimation, to generate images that better match the target style. Please check the style of the target images in references \cite{pokemonblip, lineart, nasa}.  }
       \label{fig:results_finetuning}
       \vspace{-5pt}
    \end{figure}

    We apply our pixel-domain approach from Section \ref{sec:pixel_domain} to different existing fine-tuned versions of Stable Diffusion, covering different styles and fine-tuning strategies.  Note that the fine-tuning has already been done: we do not do any additional fine-tuning with our approach. Our results for three such models are shown in Figure \ref{fig:results_finetuning}. Figure \ref{fig:results_pokemon} shows the results of the Pokemon-LoRA model \cite{pokemonLORA, hu2021lora, ryu2023low}. For this style, we use the first 50 images of the Pokemon BLIP captions dataset \cite{pokemonblip} to obtain our signal leak distribution $\sqrt{\bar{\alpha}_{T}} \tilde{\bm{x}}, \tilde{\bm{x}} \sim \tilde{q_0}$, but use all 833 images to compute the FID metrics. The results obtained with the current inference process (first row) do not correctly match the expected style, as seen qualitatively and with the FID scores. In particular, the generated images do not have a white background, as opposed to those used in training the model \cite{pokemonLORA, pokemonblip}. Sampling the initial latents with a signal leak $\sqrt{\bar{\alpha}_{T}} \Tilde{\bm{x}}$ generates images matching the expected style (second row). Similar observations are made in Figure \ref{fig:results_lineart} for the ``line-art'' style tuned with Textual Inversion \cite{gal2022image}, and in Figure \ref{fig:results_nasa} for the ``Nasa-space'' style tuned with DreamBooth \cite{ruiz2023dreambooth}. For these two styles, we use all the target images (respectively 7 and 24 images) to obtain our estimated distribution of the signal leak and to compute the FID scores. The $\text{FID}_{64}$ score is improved significantly, indicating, accordingly to the qualitative assessment, that the images generated with our approach reproduce more faithfully the style of the target images. Note that the CLIP score remains high, which implies that our approach does not affect how well the generated images match their textual prompt.
 
\subsection{Improving style for non-style-specific models}
\label{sec:notuning}   
    \begin{figure}[t]
        \begin{subfigure}[ht]{1.\linewidth}
           \centering
           \setlength{\tabcolsep}{.5pt}
            \begin{tabular}{ccccc}
        
                \makecell[c]{Original \vspace{-1mm} \\ \footnotesize $\text{FID}_{64} = 44.0$ \vspace{-1mm}\\ \footnotesize $\text{FID}_{2048} = 194$ \vspace{-1mm}\\ \footnotesize ${\mathbf{\textbf{CLIP} = 31.3}}$}  &
                \includegraphics[align=c, width=.178\linewidth]{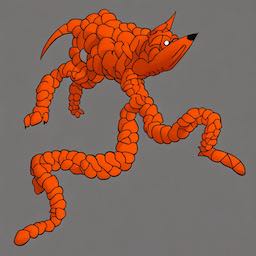} & 
                \includegraphics[align=c, width=.178\linewidth]{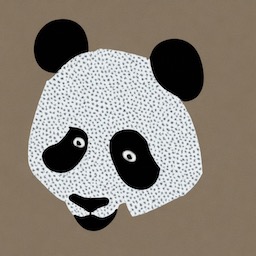} & 
                \includegraphics[align=c, width=.178\linewidth]{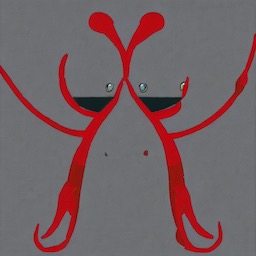} & 
                \includegraphics[align=c, width=.178\linewidth]{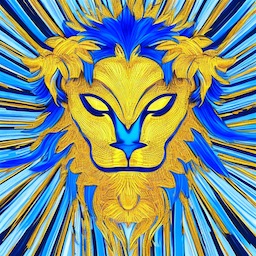} \vspace{1pt} \\
                
                \makecell[c]{Ours \vspace{-1mm} \\ \footnotesize $\mathbf{\textbf{FID}_{64} = 1.7}$ \vspace{-1mm}\\ \footnotesize $\mathbf{\textbf{FID}_{2048} = 164}$ \vspace{-1mm}\\ \footnotesize ${\text{CLIP} = 31.0}$}  &
                \includegraphics[align=c, width=.178\linewidth]{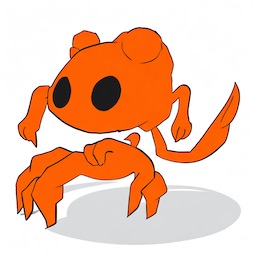} & 
                \includegraphics[align=c, width=.178\linewidth]{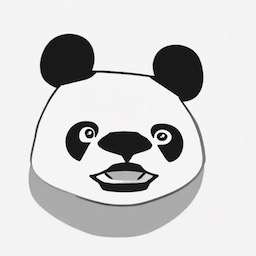} & 
                \includegraphics[align=c, width=.178\linewidth]{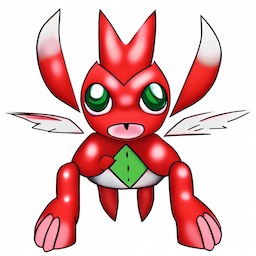} & 
                \includegraphics[align=c, width=.178\linewidth]{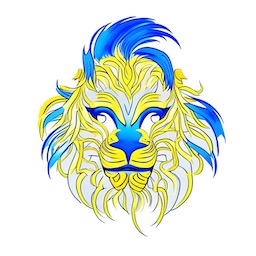} \vspace{1pt}\\
                
            \end{tabular}
           \caption{ Targeting the style of the Pokemon BLIP captions dataset \cite{pokemonblip}, generating images with the prompt ``[text] In the style of Satoshi Tajiri, white background.''}
           \label{fig:results_pokemonnotuning}
        \end{subfigure}
        
        \begin{subfigure}[ht]{1.\linewidth}
           \centering
           \setlength{\tabcolsep}{.5pt}
            \begin{tabular}{ccccc}
        
                \makecell[c]{Original \vspace{-1mm} \\ \footnotesize $\text{FID}_{64} = 105$ \vspace{-1mm}\\ \footnotesize $\text{FID}_{2048} = 423$ \vspace{-1mm}\\ \footnotesize ${\text{CLIP} = 29.6}$} &
                \includegraphics[align=c, width=.178\linewidth]{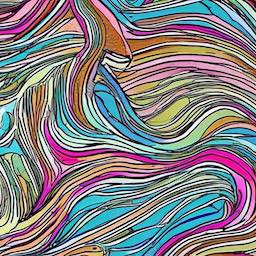} & 
                \includegraphics[align=c, width=.178\linewidth]{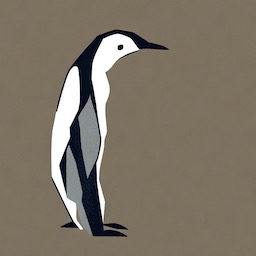} & 
                \includegraphics[align=c, width=.178\linewidth]{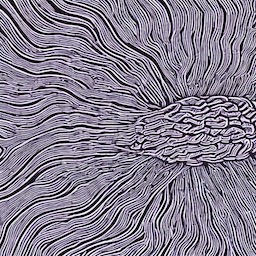} & 
                \includegraphics[align=c, width=.178\linewidth]{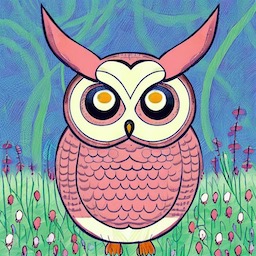} \vspace{1pt} \\

                \makecell[c]{Ours \vspace{-1mm} \\\footnotesize $\mathbf{\textbf{FID}_{64} = 5.6}$ \vspace{-1mm}\\ \footnotesize $\mathbf{\textbf{FID}_{2048} = 380}$ \vspace{-1mm}\\ \footnotesize $\mathbf{\textbf{CLIP} = 29.8}$} &
                \includegraphics[align=c, width=.178\linewidth]{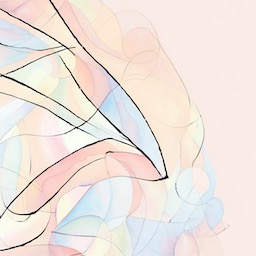} & 
                \includegraphics[align=c, width=.178\linewidth]{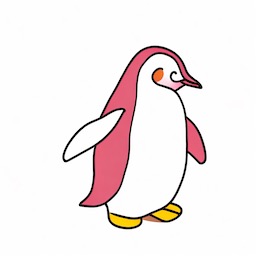} & 
                \includegraphics[align=c, width=.178\linewidth]{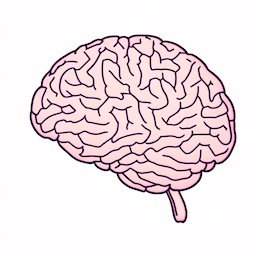} & 
                \includegraphics[align=c, width=.178\linewidth]{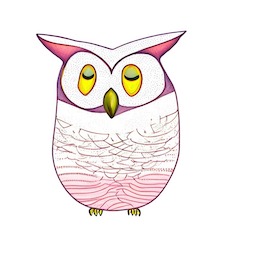} \vspace{1pt}\\
                
            \end{tabular}
           \caption{ Targeting the style of the 7 line-art images \cite{lineart}, generating images with the prompt ``[text] In the style of line art, pastel colors, white background.''}
           \label{fig:results_lineartnotuning}
        \end{subfigure}
    
        \begin{subfigure}[ht]{1.\linewidth}
           \centering
           \setlength{\tabcolsep}{.5pt}
            \begin{tabular}{ccccc}
        
                \makecell[c]{Original \vspace{-1mm} \\ \footnotesize $\text{FID}_{64} = 12.7$ \vspace{-1mm} \\ \footnotesize $\text{FID}_{2048} = 313$} &
                \includegraphics[align=c, width=.178\linewidth]{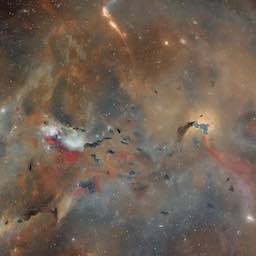} & 
                \includegraphics[align=c, width=.178\linewidth]{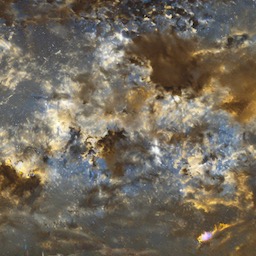} & 
                \includegraphics[align=c, width=.178\linewidth]{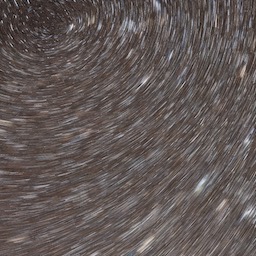} & 
                \includegraphics[align=c, width=.178\linewidth]{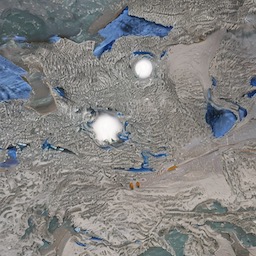} \vspace{1pt} \\
                
                \makecell[c]{Ours \vspace{-1mm} \\ \footnotesize $\mathbf{\textbf{FID}_{64} = 7.7}$ \vspace{-1mm} \\ \footnotesize $\mathbf{\textbf{FID}_{2048} = 223}$} &
                \includegraphics[align=c, width=.178\linewidth]{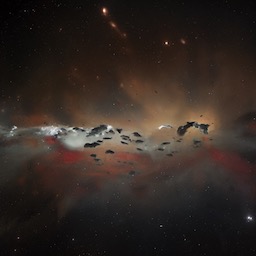} & 
                \includegraphics[align=c, width=.178\linewidth]{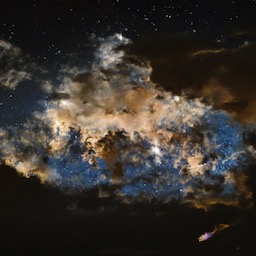} & 
                \includegraphics[align=c, width=.178\linewidth]{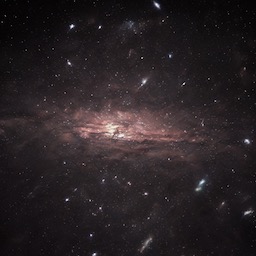} & 
                \includegraphics[align=c, width=.178\linewidth]{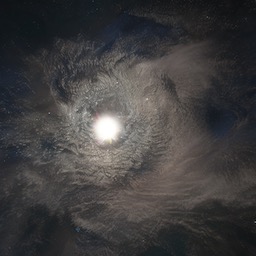} \vspace{1pt}\\
                
            \end{tabular}
           \caption{Targeting the style of the 24 images of astronomical phenomena \cite{nasa}, generating images with the prompt ``A photo of the sky made by the NASA.''}
           \label{fig:results_nasanotuning}
        \end{subfigure}
       \caption{When using white noise as initial latent, describing the style in the textual prompt is insufficient to generate images that match the desired style, as illustrated in the first row of each subfigure \ref{fig:results_pokemonnotuning}, \ref{fig:results_lineartnotuning} and \ref{fig:results_nasanotuning}. In each of the second rows, images are generated with the same prompts and model using our proposed approach, sampling a signal leak $\sqrt{\bar{\alpha}_{T}} \tilde{\bm{x}}$ from our pixel-domain estimation. We used the same approach as in Section \ref{sec:finetuning} to estimate the distribution in the pixel domain for the three styles. All images here are generated with standard Stable Diffusion 2.1, without fine-tuning for specific styles. }
       \label{fig:results_notuning}
    \end{figure}

    We additionally experiment with exploiting the signal-leak bias directly in the original diffusion model, without using a tuned version of the model. 
    Simply describing a desired style in the textual prompt is often insufficient to generate images in the desired style \cite{diffusion_in_style}. However, when we combine the style description with our approach of exploiting the signal-leak bias, the generated images seem to match well the desired style. This suggests that fine-tuning the model for a specific style may not always be necessary, as shown in Figure \ref{fig:results_notuning}.
    It is possible to generate images in a particular style by exploiting the signal-leak bias without any fine-tuning. By putting a signal leak $\sqrt{\bar{\alpha}_{T}} \tilde{\bm{x}}, \tilde{\bm{x}} \sim \tilde{q_0}$ into the initial latent, we bias the denoising process toward generating images that look like $\tilde{\bm{x}}$. Our strategy here takes only a few seconds to estimate the distribution $\tilde{q_0}$, all without any compromise on the inference time to generate an image, for instance as opposed to guidance \cite{pan2023arbitrary}. \textbf{Limitation:} Some specific styles may not be easily described in words and may correspond to characteristics not captured by our pixel-domain model. We design such an example in Section 2.4 of the supplementary material. For such styles, fine-tuning or a different model for $\tilde{q_0}$ would be required.
    
\subsection{Generating more varied images}
\label{sec:varied_images}

\begin{figure}[h]
       \centering
       \setlength{\tabcolsep}{.1pt}
        \begin{tabular}{cccccccc}    
            \multicolumn{8}{c}{\makecell[c]{Original \\ \footnotesize $\text{FID}_{64} = 2.65, \quad{\text{FID}_{2048} = 192, \quad \mathbf{\textbf{CLIP} = 32.2}}$}} \\
            \includegraphics[align=c, width=.124\linewidth]{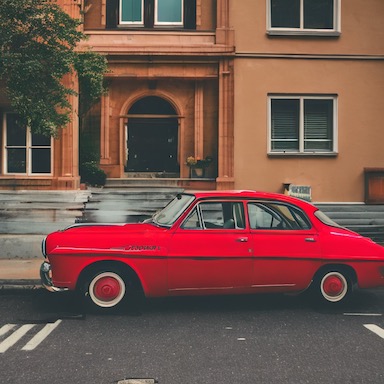} & 
            \includegraphics[align=c, width=.124\linewidth]{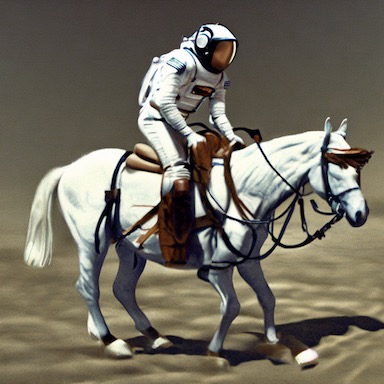} & 
            \includegraphics[align=c, width=.124\linewidth]{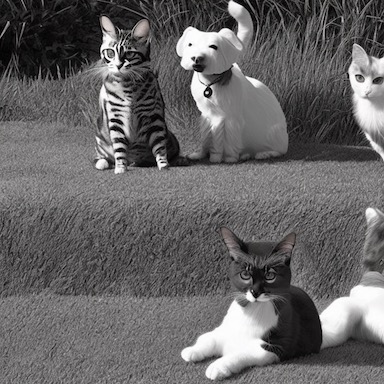} & 
            \includegraphics[align=c, width=.124\linewidth]{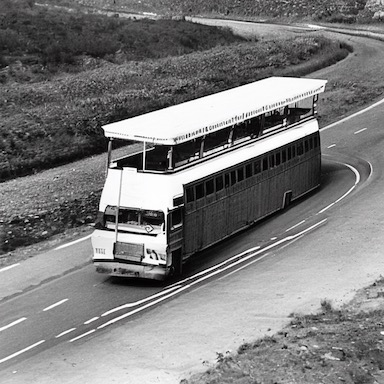} & 
            \includegraphics[align=c, width=.124\linewidth]{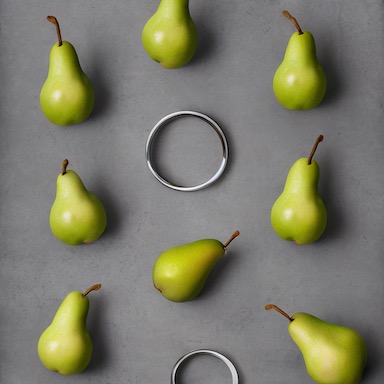} & 
            \includegraphics[align=c, width=.124\linewidth]{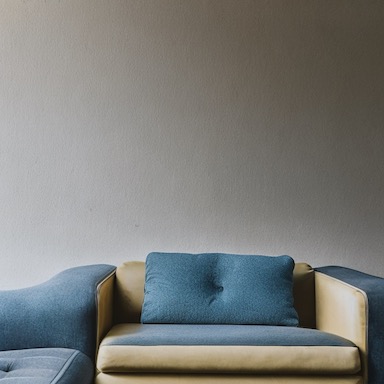} & 
            \includegraphics[align=c, width=.124\linewidth]{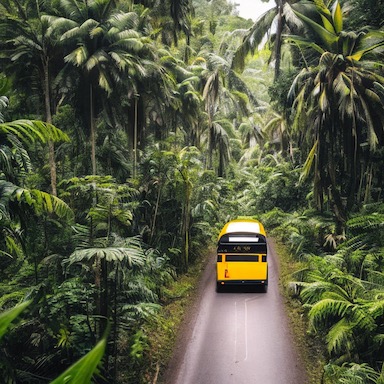} & 
            \includegraphics[align=c, width=.124\linewidth]{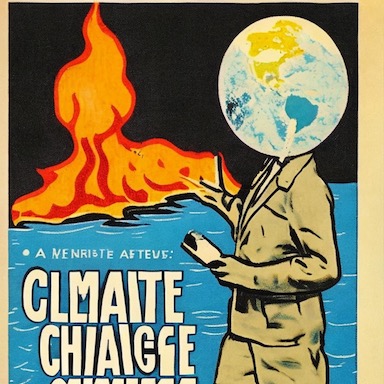} 
            \vspace{.1pt} \\
            
            \includegraphics[align=c, width=.124\linewidth]{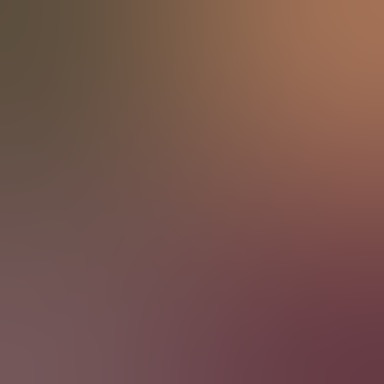} & 
            \includegraphics[align=c, width=.124\linewidth]{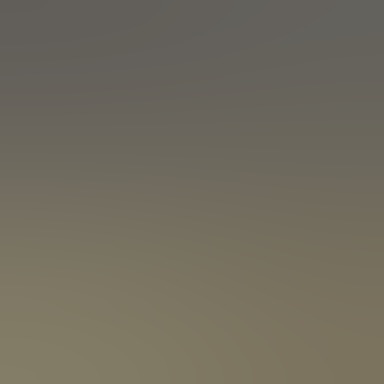} & 
            \includegraphics[align=c, width=.124\linewidth]{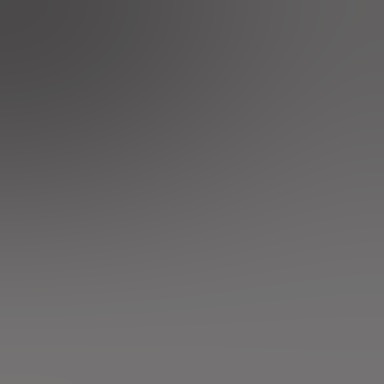} & 
            \includegraphics[align=c, width=.124\linewidth]{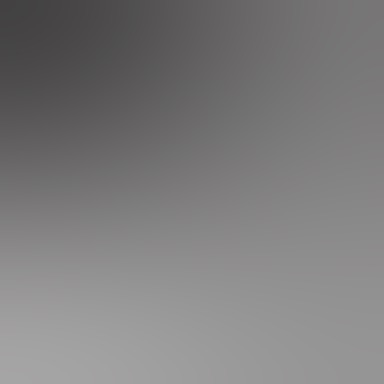} & 
            \includegraphics[align=c, width=.124\linewidth]{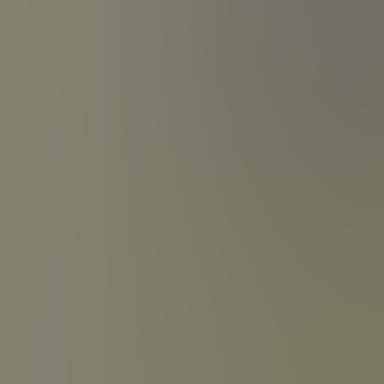} & 
            \includegraphics[align=c, width=.124\linewidth]{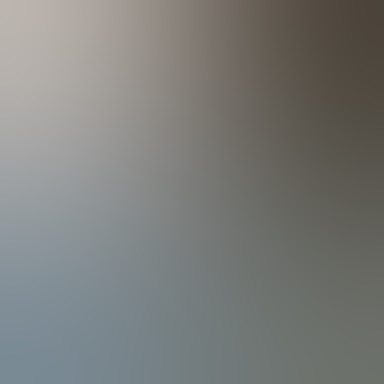} & 
            \includegraphics[align=c, width=.124\linewidth]{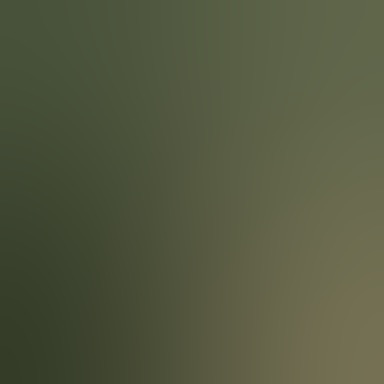} & 
            \includegraphics[align=c, width=.124\linewidth]{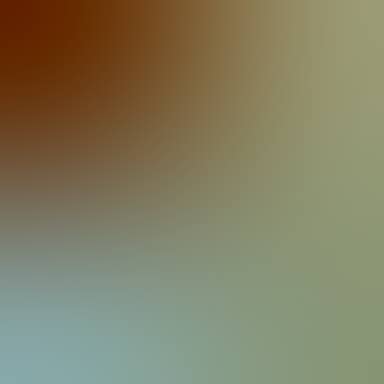} 
            \vspace{3pt} \\

            \multicolumn{8}{c}{\makecell[c]{Ours \\ \footnotesize $\mathbf{\textbf{FID}_{64} = 2.64}, \quad\mathbf{\textbf{FID}_{2048} = 187}, \quad \text{CLIP} = 31.8$}} \\
            \includegraphics[align=c, width=.124\linewidth]{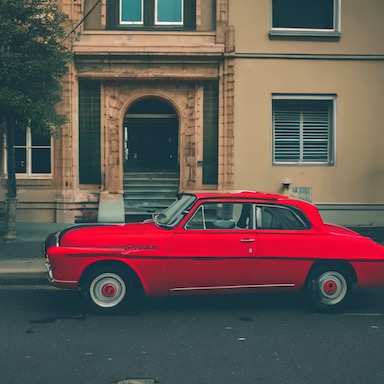} & 
            \includegraphics[align=c, width=.124\linewidth]{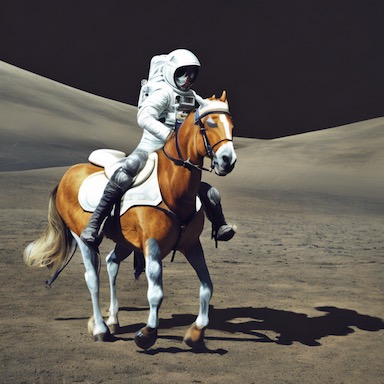} & 
            \includegraphics[align=c, width=.124\linewidth]{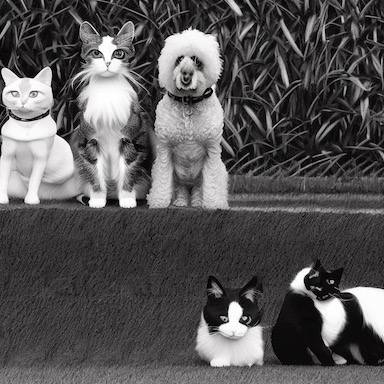} & 
            \includegraphics[align=c, width=.124\linewidth]{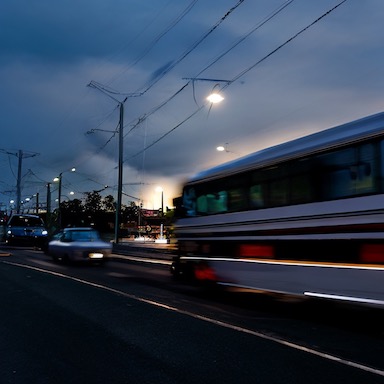} & 
            \includegraphics[align=c, width=.124\linewidth]{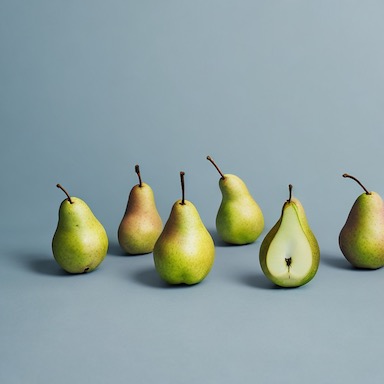} & 
            \includegraphics[align=c, width=.124\linewidth]{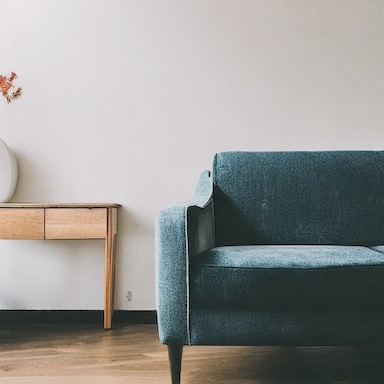} & 
            \includegraphics[align=c, width=.124\linewidth]{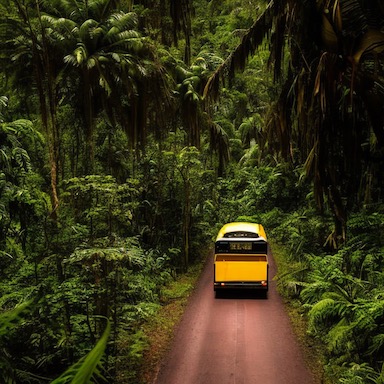} & 
            \includegraphics[align=c, width=.124\linewidth]{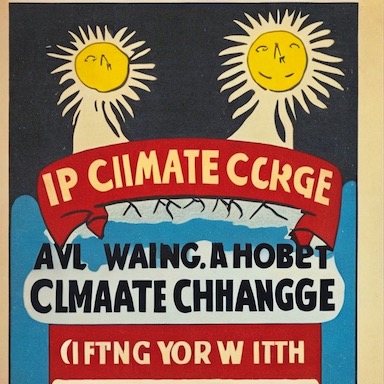} 
            \vspace{.1pt} \\
             
            \includegraphics[align=c, width=.124\linewidth]{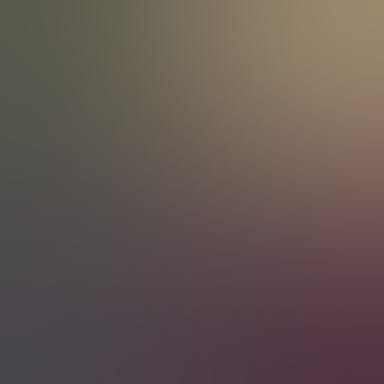} & 
            \includegraphics[align=c, width=.124\linewidth]{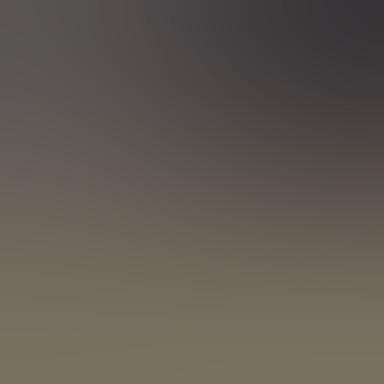} & 
            \includegraphics[align=c, width=.124\linewidth]{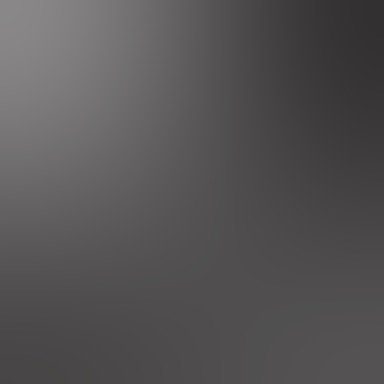} & 
            \includegraphics[align=c, width=.124\linewidth]{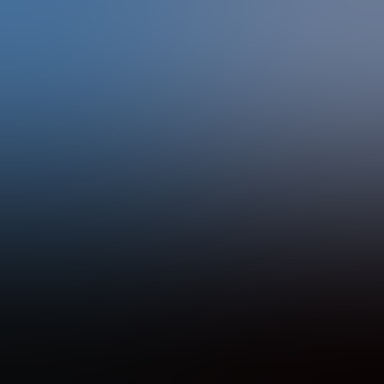} & 
            \includegraphics[align=c, width=.124\linewidth]{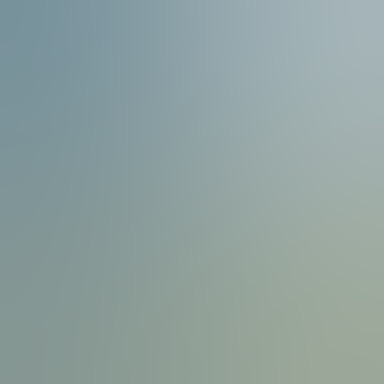} & 
            \includegraphics[align=c, width=.124\linewidth]{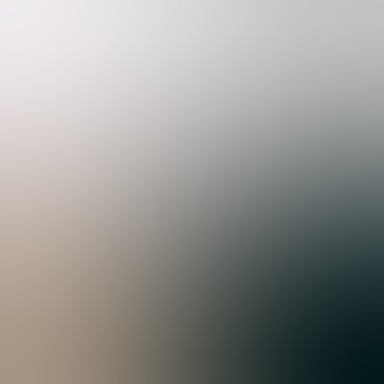} & 
            \includegraphics[align=c, width=.124\linewidth]{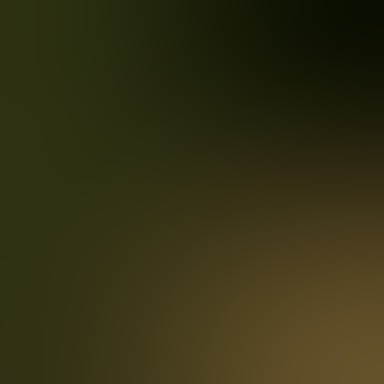} & 
            \includegraphics[align=c, width=.124\linewidth]{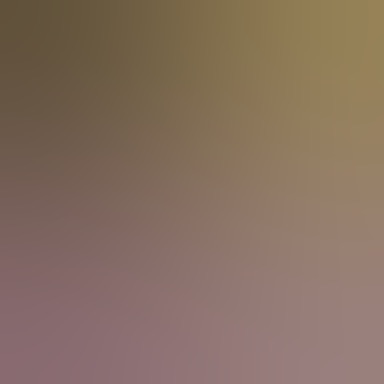} 
            \vspace{0pt} \\            
        \end{tabular}
       \caption{We generated 8 images from Stable Diffusion 2.1, at the default resolution $768 \times 768$ with 8 prompts from DrawBench \cite{saharia2022photorealistic}, in the top starting from white noise (\ie the default behavior), and in the bottom from our distribution realigned with the distribution of natural images, as explained in Sections \ref{sec:freq_domain} and \ref{sec:varied_images}. Below each generated image, we show a low-pass filtered version of it. We observe, as expected from our analysis, that images generated by taking into account the signal leak have more varied low-frequency components, as in natural images.
       Images in the top tend to be greyish, with medium brightness and little variation of colors inside each image.
       Images in the bottom have more varied colors and brightness across images. They tend to be less greyish and to have more variation of colors inside each image.
       This all comes without any additional training of Stable Diffusion.  Quantitative comparisons of average pixel values, contrast, and luminance are provided in the supplementary Section 3.2.
       }
       \label{fig:results_varied}
    \end{figure}

    As mentioned earlier, images currently generated with Stable Diffusion tend to have medium low-frequency components, \eg medium brightness and little variation of colors between different areas of an image. This observation is noticeable in the top rows of Figure \ref{fig:results_varied}. To generate images with more varied low-frequency components, we apply our approach from Section \ref{sec:freq_domain} by estimating the distribution of the signal leak in both frequency and pixel domains. Especially, we use 323 images from the LAION-6+ dataset \cite{6plus} to model the 3 lowest-frequency components, \ie a value $N=3$ following the notation in Section \ref{sec:freq_domain}. We use the same 323 images to compute the FID scores.

    We visualize in Figure \ref{fig:results_varied} the effect of using our method to sample the initial latents instead of sampling them from white noise. The effect is slight, but noticeable on the 8 randomly-picked images of this figure. Images generated by sampling from a distribution containing a signal leak with more varied low-frequency components also have more varied low-frequency components. This not only solves the issue of generating ``only'' medium-brightness images but, also results in more natural variances of colors inside each image; this all without extra training, as opposed to previous solutions \cite{guttenberg2023diffusion, lin2023common}. 
    
    The FID scores are slightly improved, suggesting, according to our visual assessment, that images better match the distribution of low-level features of natural images. The $\text{CLIP}$ score worsens only very slightly, suggesting our approach has almost no impact on the content and high-frequency alignment of the generated images.     
    \textbf{Limitation:} The signal leak $\sqrt{\bar{\alpha}_{T}} \tilde{\bm{x}}, \tilde{\bm{x}} \sim \tilde{q_0}$ is sampled randomly with our approach. One advantage of prior work based on retraining \cite{guttenberg2023diffusion, lin2023common} is that the brightness of the generated image matches the textual prompt instead of being random.

\subsection{Explicit influence on low-frequency attributes}
\label{sec:control}

\begin{figure}[t]
       \centering
       \setlength{\tabcolsep}{.5pt}
        \begin{tabular}{ccc c ccc}
    
            \includegraphics[align=c, width=.13\linewidth]{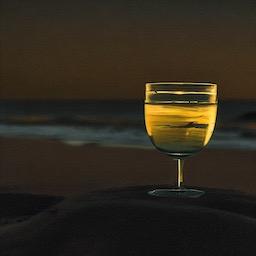} & 
            \includegraphics[align=c, width=.13\linewidth]{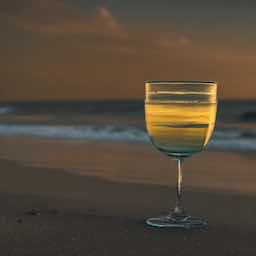} & 
            \includegraphics[align=c, width=.13\linewidth]{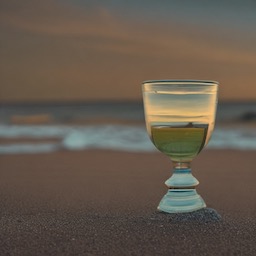} & 
            \includegraphics[align=c, width=.13\linewidth]{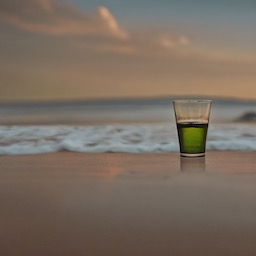} &
            \includegraphics[align=c, width=.13\linewidth]{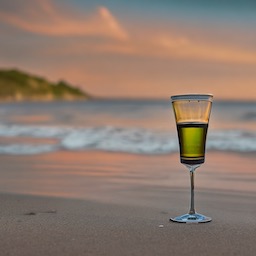} & 
            \includegraphics[align=c, width=.13\linewidth]{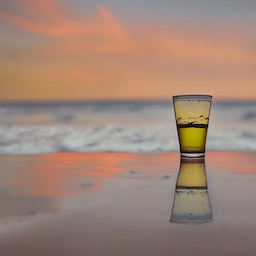} & 
            \includegraphics[align=c, width=.13\linewidth]{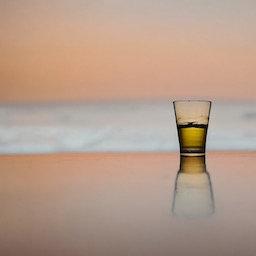} \vspace{1pt} \\
            \includegraphics[align=c, width=.13\linewidth]{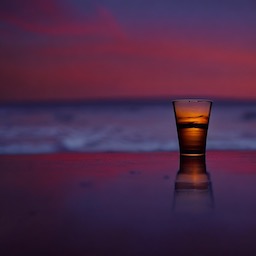} & 
            \includegraphics[align=c, width=.13\linewidth]{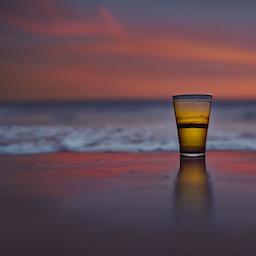} & 
            \includegraphics[align=c, width=.13\linewidth]{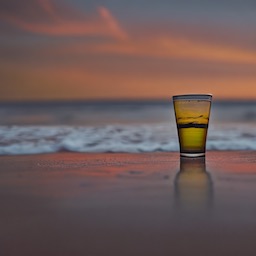} & 
            \includegraphics[align=c, width=.13\linewidth]{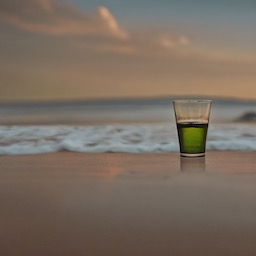} &
            \includegraphics[align=c, width=.13\linewidth]{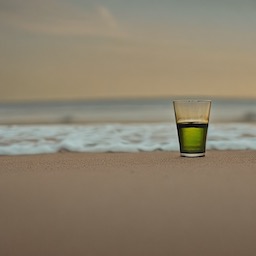} & 
            \includegraphics[align=c, width=.13\linewidth]{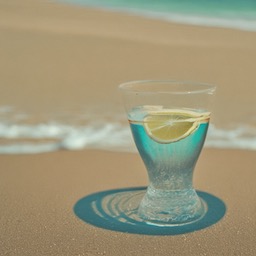} & 
            \includegraphics[align=c, width=.13\linewidth]{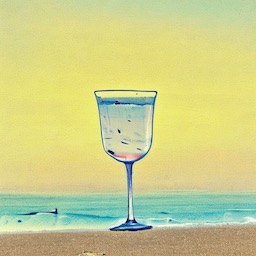} \vspace{1pt} \\
            \includegraphics[align=c, width=.13\linewidth]{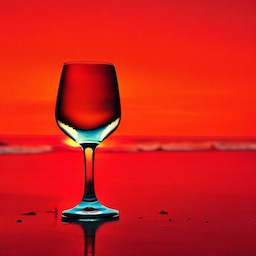} & 
            \includegraphics[align=c, width=.13\linewidth]{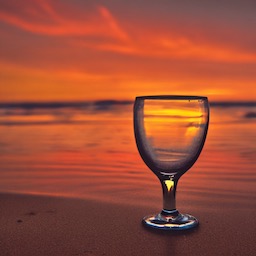} & 
            \includegraphics[align=c, width=.13\linewidth]{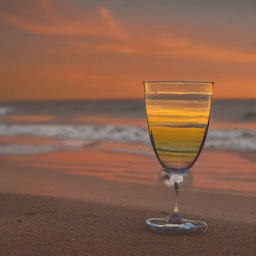} & 
            \includegraphics[align=c, width=.13\linewidth]{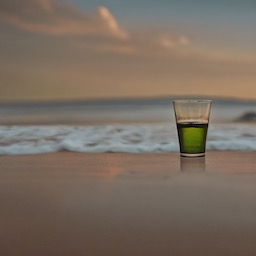} &
            \includegraphics[align=c, width=.13\linewidth]{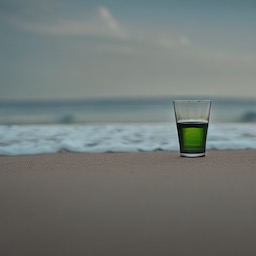} & 
            \includegraphics[align=c, width=.13\linewidth]{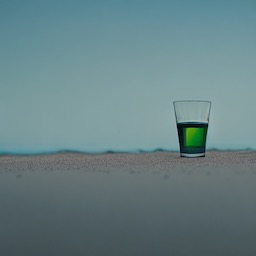} & 
            \includegraphics[align=c, width=.13\linewidth]{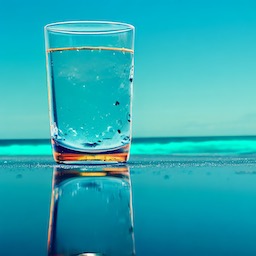} \vspace{1pt} \\
            \includegraphics[align=c, width=.13\linewidth]{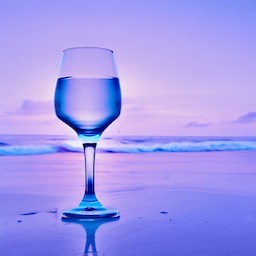} & 
            \includegraphics[align=c, width=.13\linewidth]{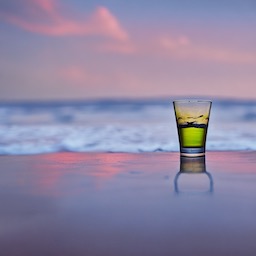} & 
            \includegraphics[align=c, width=.13\linewidth]{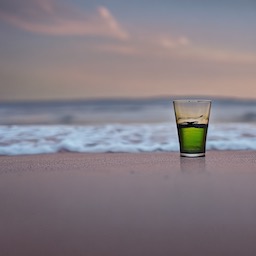} & 
            \includegraphics[align=c, width=.13\linewidth]{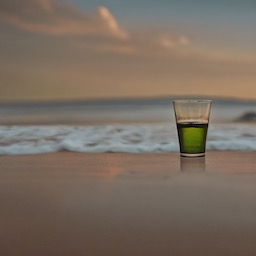} &
            \includegraphics[align=c, width=.13\linewidth]{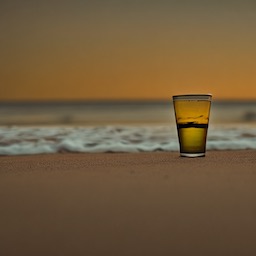} & 
            \includegraphics[align=c, width=.13\linewidth]{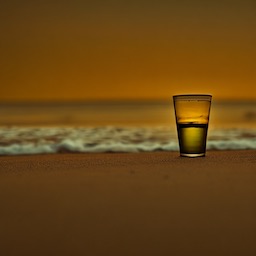} & 
            \includegraphics[align=c, width=.13\linewidth]{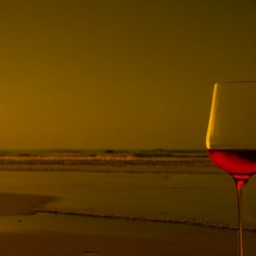} \vspace{1pt} \\
            -2.0 & -1.0 & -0.5 & 0 & 0.5 & 1.0 & 2.0
            
        \end{tabular}
       \caption{Explicit control on the mean color of the generated images. Instead of sampling the signal leak $\sqrt{\bar{\alpha}_{T}} \tilde{\bm{x}}$ from the computed distribution, we can manually set its low-frequency components $\Tilde{\bm{X}}_{\text{LF}}$ and randomly sample only the high-frequency components $\Tilde{\bm{x}}_{\text{HF}}\sim \Tilde{q_{0}}_\text{HF}$. This provides a specific bias towards generating images with desired low-frequency components, here, the mean color of the image. In this figure, images are generated from the prompt ``A glass on the beach'' with Stable Diffusion 2.1 by setting $\Tilde{\bm{X}}_{\text{LF}} \in \mathbb{R}^{4}$ to $\bm{0}$, except for one of the four channels, for which we set the value indicated at the bottom of each column. The four rows correspond to the 4 channels in Stable Diffusion's latent space. }
       \label{fig:results_control}
    \end{figure}

    Instead of randomly sampling the signal leak $\sqrt{\bar{\alpha}_{T}} \Tilde{\bm{x}}$, its low-frequencies components can be manually selected by the user. This provides explicit control over the generated image atop the textual prompt, without needing any target images.   
    Following the notations of Section \ref{sec:freq_domain}, we set $N=1$ and use 323 images from the LAION-6+ dataset~\cite{6plus} to obtain $\Tilde{q_{0}}_\text{HF}$. Instead of sampling $\Tilde{\bm{x}}$ as in Equation \ref{eq:fred_new_dist}, we manually select a value for $\Tilde{\bm{X}}_\text{LF}$ and sample $\Tilde{\bm{x}}$ as $\text{IDCT}(\Tilde{\bm{X}}_\text{LF})+\Tilde{\bm{x}}_\text{HF}$, with $\Tilde{\bm{x}}_\text{HF} \sim \Tilde{q_{0}}_\text{HF}$. As we show in Figure \ref{fig:results_control}, it is easy to interpret the effect of the different values of $\Tilde{\bm{X}}_\text{LF}$. We can consistently bias the generation of images towards a specific brightness or desired colors.

\section{Conclusion}

    In this paper, we show that the signal-leak bias in diffusion models is not only caused by a non-zero SNR during the training of the last timestep, but also a discrepancy between the noise and the data distributions. 
    When generating natural images, the discrepancy between the noise and the data distributions lies in the frequency domain, explaining why generated images always tend to have medium low-frequency values, including medium brightness.
    When diffusion models are tuned to a specific style, the discrepancy between the noise and the data distributions lies in the pixel domain, explaining the unsatisfactory outcomes of style adaptation of diffusion models.  
    
    We propose a simple way to exploit this signal-leak bias to our advantage to solve these issues. By injecting a signal leak in the initial latent at inference time, we can bias the image generation toward a desired specific color distribution or a specific style. This simple step does not require any fine-tuning making it much simpler than existing approaches for style or color-specific image generation. 
    
    We encourage future research to account for training and inference distribution gap when training or fine-tuning diffusion models, and to include a signal leak in the initial latents at inference time as well, in order to mirror the training process and achieve visually more pleasing results.  

    \textbf{Acknowledgements:} This work was supported by Innosuisse grant 48552.1 IP-ICT.

{\small
\bibliographystyle{ieee_fullname}
\bibliography{main.bib}
}

\end{document}